\documentclass{article} 
\usepackage{colm2024_conference}

\usepackage{hyperref}
\usepackage{multirow}
\usepackage{graphicx}
\usepackage{makecell}
\usepackage{pdfpages}
\usepackage{url}
\usepackage{xcolor}
\usepackage{epsfig}
\usepackage{xspace}
\usepackage{adjustbox}
\usepackage{amsfonts}
\usepackage{amsmath}
\usepackage{amssymb}
\usepackage{booktabs} 
\usepackage{comment}
\usepackage{caption}
\usepackage{subcaption}
\usepackage{textcomp}
\usepackage{relsize}
\usepackage{stmaryrd}
\usepackage{bbm}
\usepackage{rotating}
\usepackage{helvet}
\usepackage{courier}
\usepackage{natbib}
\usepackage{cleveref}
\usepackage{array}
\usepackage{lipsum}
\usepackage{amsmath}
\usepackage{microtype}
\usepackage{hyperref}
\usepackage{algorithm}
\usepackage{algpseudocode}
\usepackage{algorithmicx}
\usepackage{url}
\usepackage{hyperref}
\usepackage{multirow}
\usepackage{graphicx}
\usepackage{pdfpages}
\usepackage{url}
\usepackage{xcolor}
\usepackage{epsfig}
\usepackage{adjustbox}
\usepackage{amsfonts}
\usepackage{amsmath}
\usepackage{amssymb}
\usepackage{booktabs} 
\usepackage{comment}
\usepackage{caption}
\usepackage{subcaption}
\usepackage{textcomp}
\usepackage{relsize}
\usepackage{stmaryrd}
\usepackage{bbm}
\usepackage{rotating}
\usepackage{helvet}
\usepackage{courier}
\usepackage{natbib}
\usepackage{cleveref}
\usepackage{array}
\usepackage{booktabs}
\usepackage{graphicx} 
\usepackage{enumitem}
\definecolor{darkblue}{rgb}{0, 0, 0.5}
\hypersetup{colorlinks=true, citecolor=darkblue, linkcolor=darkblue, urlcolor=darkblue}
\usepackage[switch]{lineno}

\title{Ever-Evolving Memory by Blending and Refining the Past}

\author{Seo Hyun Kim$^1$, Keummin Ka$^1$, \\
\bf{Yohan Jo$^2$, Seung-won Hwang$^2$, Dongha Lee$^1$, Jinyoung Yeo$^1$} \\
\\
\bf{$^1$Yonsei University, $^2$Seoul National University} \\
\texttt{\small $^1$\{shkimsally, kummin0429, donalee, jinyeo\}@yonsei.ac.kr}\\
\texttt{\small $^2$\{yohan.jo, seungwonh\}@snu.ac.kr}
}

\colmfinalcopy 

\begin{document}

\maketitle

\begin{abstract}
For a human-like chatbot, constructing a long-term memory is crucial. However, current large language models often lack this capability, leading to instances of missing important user information or redundantly asking for the same information, thereby diminishing conversation quality. To effectively construct memory, it is crucial to seamlessly connect past and present information, while also possessing the ability to forget obstructive information. To address these challenges, we propose \textsc{CREEM}, a novel memory system for long-term conversation. Improving upon existing approaches that construct memory based solely on current sessions, \textsc{CREEM} blends past memories during memory formation. Additionally, we introduce a refining process to handle redundant or outdated information. Unlike traditional paradigms, we view responding and memory construction as inseparable tasks. The blending process, which creates new memories, also serves as a reasoning step for response generation by informing the connection between past and present. Through evaluation, we demonstrate that \textsc{CREEM} enhances both memory and response qualities in multi-session personalized dialogues. 
\end{abstract}
\section{Introduction}
To enhance the conversational experience with human-like chatbots, chatbots should possess long-term memory and utilize it to engage users in conversations. While today's large language models (LLMs) have facilitated interactions with humans, they often lack long-term memory. This limitation leads to instances of forgetting important user information or redundantly exchanging information, ultimately diminishing the quality of conversation~\citep{zhong2024memorybank}. To bridge this gap and provide a more authentic conversational experience, constructing and integrating long-term memory into chatbot is imperative. 

In long-term conversations, the user's situation and preference can continue to change as the number of sessions increases. For instance, one who was into classical music might turn into a passionate fan of heavy metal, or one who was suffering from severe illness could now be healthy enough to go jogging every morning. In order to create a chatbot that is sensitive to these changes, it is crucial that the changes are immediately reflected in its long-term memory~\citep{adiwardana2020towards}. 

\begin{figure*}[t] 
    \begin{center}
    \includegraphics[width=1.0\linewidth]{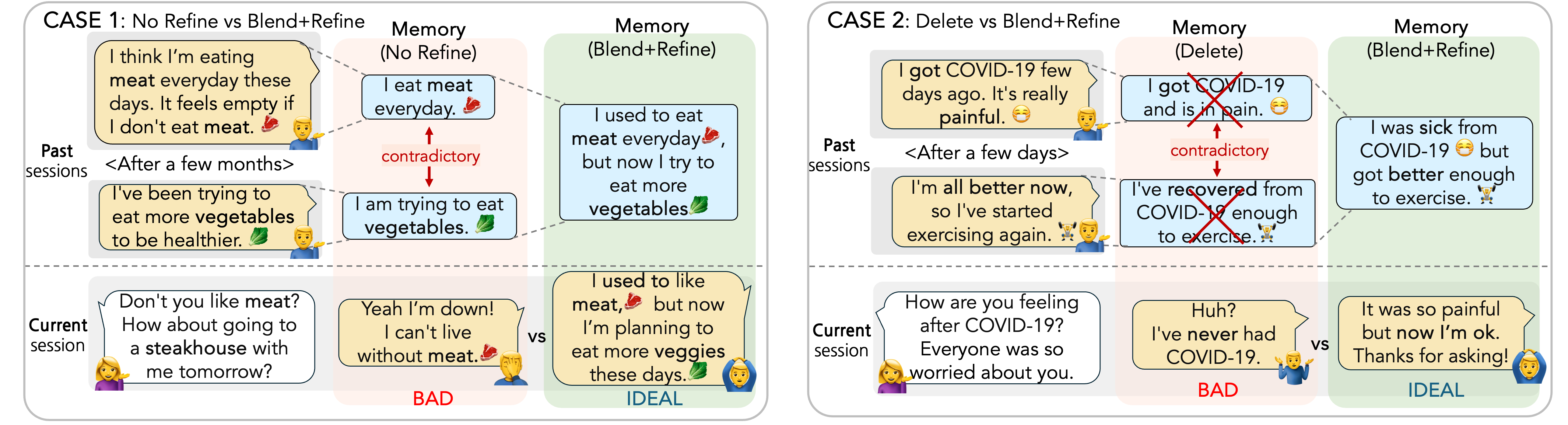}
    \end{center}
    \caption{Motivating examples where our memory
(Blend + Refine) is needed compared to not refining
(Case 1) and deleting all contradictories (Case 2).}
    \label{motivating_examples}
\end{figure*}

When constructing a memory system, the approach of simply storing the summarized dialogue or extracting persona may seem intuitive~\citep{jang2023conversation, xu2021beyond}. However, these approaches do not effectively reflect changes in the conversation. They focus solely on the current session when extracting information to save, resulting in a listing of isolated information without capturing changes between sessions. Furthermore, as memory accumulates without a refining process, outdated or contradictory information may be intermixed~\citep{bae2022keep}. This hinders the retrieval of correct information, potentially leading to inappropriate responses as shown in Case 1 of Figure ~\ref{motivating_examples}. Therefore, any memories that become useless or obstructive should be deliberately forgotten. While the refinement of memory is essential, research in this area remains insufficiently explored. To address both aspects effectively, we focus on the following question: \textit{How can we store past and new information without causing confusion or interference between them?}




To this end, we propose a novel memory framework called \textbf{C}ontextualized \textbf{R}efinement based \textbf{E}ver-\textbf{E}volving \textbf{M}emory, \textsc{CREEM}. \textsc{CREEM} blends past and present information to generate insights, which are then added as new memories. Subsequently, refining operations are employed to forget outdated or redundant past memories. In contrast to traditional paradigms, we do not view response generation and memory construction as separate tasks. Our blending process, where new insights are created, serves to link the past and present. The insights inferred in this process are then infused to the response generation model, assisting in properly integrating the updated information. Therefore, the blending process serves as both a step for memory construction and reasoning for response generation.

To demonstrate the effectiveness of \textsc{CREEM}, we begin by conducting memory evaluation using G-Eval along with introducing new criteria. Additionally, we closely examine the impact of refining by assessing the level of contradiction using a natural language inference (NLI) model, and through questions that address changes in content across sessions. For response generation evaluation, we conduct both reference-based and reference-free evaluations. We observe that even without fine-tuning, our model displayed strong performance in memory augmented response generation.

In summary, our main contributions are as follows:

\begin{itemize}
    \item We introduce \textsc{CREEM}, a novel memory constructing framework to tackle the issues of fragmented and contradictory memories.
    \item We enhance memory adapted responding ability by infusing the blended insight to the response generation model.
    \item We establish new criteria for what constitutes a good memory and evaluate based on them.
    \item We demonstrate how the memory quality aligns with the response quality.
\end{itemize}

\section{Related Work}
\subsection{Long-term Memory Systems}
The long-term memory system has been an ongoing subject of research. \citet{xu2021beyond} extracts persona by summarizing the utterances for each speakers and use them as a memory for generating response in ongoing conversation. They introduce Multi-Session Chat (MSC), a dataset created by human annotators through expanding dialogues from Persona-Chat~\citep{zhang2018personalizing}
In \citet{jang2023conversation}, memory is constructed by summarizing the each session in order to remember the content of the previous session and have a conversation in the next session. 
They introduce a new dataset, Conversational Chronicles (CC), which is generated by ChatGPT given a relationship, time interval, and a storyline employing the narratives from SODA~\citep{kim2022soda}.
These methods of building memory by summarizing conversations creates one-dimensional information, which may prevent high-level conversation or conversation appropriate to the situation. Meanwhile, our work creates new memories by integrating relevant past memories, rather than simply summarizing the current conversation. 
Research on more advanced versions of memory systems has also been conducted. \citet{majumder2020like} expands the user's persona with the commonsense knowledge graph~\citep{kim2023persona}. In \citet{park2023generative}, its generative agent conducts high-level reflections based on observations of the world, and then add them to the memory. Motivated by this approach, we use GPT-3.5-turbo~\citep{openai2023chatgpt} to extract insights from the current dialogue to get high-level information.

\subsection{Forgetting in Long-term Memory}
When constructing long term memory, as sessions become longer, unnecessary or conflicting information may accumulate. Therefore, there is a need for a function in memory to forget information accordingly. \citet{bae2022keep} propose a long-term memory management mechanism with memory refining operations. The operations remove or add memory by determining the relationship between past memory and the summary created from the current session. Among them, the 'delete' operation deletes both memory when they are no longer needed to be remembered and the summary contains any content that is no longer true. Case 2 in Figure~\ref{motivating_examples} shows how the deleting operation can affect the memory capability. We implement its mechanism to compare with our method through further evaluations. \citet{zhong2023memorybank} introduce a memory updating mechanism, mirroring the Ebbinghaus Forgetting Curve theory. It selectively forgets and strengthens memories based on time elapsed and importance. However it does not handle contradiction in memory.

\begin{figure*}[t] 
    \begin{center}
    \includegraphics[width=1.0\linewidth]{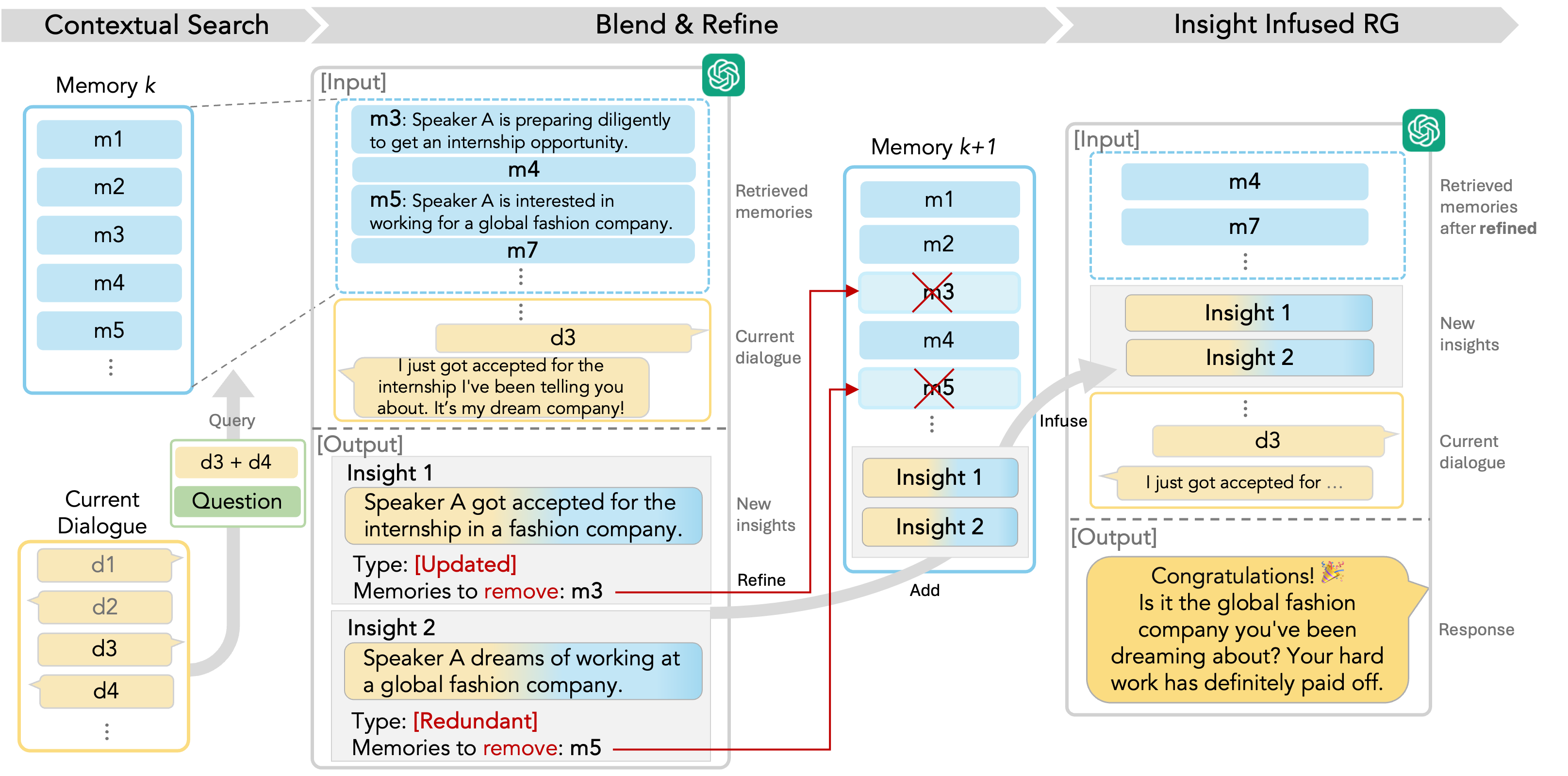}
    \end{center}
    \caption{Overall memory update process of \textsc{CREEM}.}
    \label{Overall Method}
\end{figure*} 

\section{Proposed Approach: \textsc{CREEM}}



In this section, we explain the details of each step that is executed every turn from constructing memory to generating responses. The overall method flow is shown in Figure~\ref{Overall Method} and Algorithm~\ref{alg:mem}.

\subsection{Contextual Search} \label{subsection:contextual_search}
The initial step involves contextual searching to gather a robust pool of past memories, which serve as a valuable source for blending when forming new ones. Typically, a dialogue context is utilized as a query to retrieve pertinent memories. We take a step further by generating a question that aligns with the current dialogue context and using it as an additional query. This approach enables us to retrieve a broader range of memories that are implicitly related, extending beyond those directly resembling the ongoing dialogue. We utilize GPT-3.5-turbo~\citep{openai2023chatgpt}(for short ChatGPT) for this process.

\subsection{Blend and Refine} \label{subsection:blend_refine}
\paragraph{Blending.} When creating a new memory, past memories are taken into consideration to ensure coherence and relevance. The \textbf{Blend} process uses the top-k information that were previously retrieved to generate $n$ memories through integration and high-level inference. Prior works that construct memory with the information extracted from a single session can be expressed as equation \ref{eqn:prior_eq} where $\mathrm{m}_s$ and $\mathbf{D}^s$ are new memories and dialogue at current time $s$ respectively, and $f$ is the memory expanding function. On the contrary, our method utilizes past memories while constructing the memory itself as equation \ref{eqn:our_eq} where $\mathbf{M}_{[1,s-1]}$ is the set of past memories accumulated.
\begin{align} \label{eqn:prior_eq}
	\mathrm{m}_s = f(\mathbf{D}^s)\\ \label{eqn:our_eq}
	 \mathrm{m}_s = f(\mathbf{D}^s, \mathbf{M}_{[1,s-1]})
\end{align}

\paragraph{Refining.} When adding new memory, past memory should be refined to match the overall consistency. During the \textbf{Refine} step, the relationship between new and past memories is evaluated and categorized into three types: \textit{New} if it introduces entirely novel information, \textit{Redundant} if it overlaps with existing memories, and \textit{Updated} if past information has been updated. Subsequently, based on the type of insight, appropriate actions are taken regarding past information. 
In the case of \textit{Redundant} or \textit{Updated}, LLM identifies which past memories to remove. After deletion, the new insights are added to the memory stream.

\subsection{Insight Infused Response Generation} \label{subsection:response_generation}
To generate responses, we provide the ongoing dialogue and the retrieved past memories that have undergone the refining process, along with newly created insights. The generated insights reflect changes or supplemented information based on the current utterance, thereby encapsulating connections between past and present. Through those insights, we enable the response generation model to comprehend the connections that embody changes between past and present. This ensures that responses accurately reflect the changes without any inconsistencies.

\begin{algorithm}
\begin{algorithmic}[1]
\small
\caption{Memory Updating Process}
\label{alg:mem}
\State \textbf{\fontsize{10pt}{12pt} Definition}: 
{\fontsize{10pt}{12pt} $\mathbf{M} = \textit{set of entire memories}$}, 
{\fontsize{10pt}{12pt} $\mathbf{D} = \textit{set of dialogue utterances}$}, 
\Statex
{\fontsize{10pt}{12pt} $s = \textit{session number}$}, 
{\fontsize{10pt}{12pt} $t = \textit{turn number}$}, 
\Statex
{\fontsize{10pt}{12pt} $\mathbf{M}_t^s = \textit{set of memories created at turn t of session s}$}, 
{\fontsize{10pt}{12pt} $\mathbf{M}_{remove} = \textit{set of memories to remove}$}

\State \textbf{\fontsize{10pt}{12pt} Input:} {\fontsize{10pt}{12pt}M, D , \textit{s} , \textit{t}}
\State {\fontsize{10pt}{12pt} $query \gets GenerateQuery \left(\mathbf{D}^s_{[1,t-1]}\right)$}
\State {\fontsize{10pt}{12pt} $\mathbf{[\textit{m'},...]} \gets Retrieval\left(\mathbf{M}_{[1,s-1]}|query, \mathbf{D}^s_{[t-2,t-1]}\right)$}
\State {\fontsize{10pt}{12pt} $\mathbf{M}_t^s, type, \mathbf{M}_{remove} \gets Blend\&Refine(\mathbf{[\textit{m'},...]},\mathbf{D}^s_{[1,t-1]})$ }

\State $Response \gets GenerateResponse\left(\mathbf{M}_t^s, \mathbf{[\textit{m'},...]}_{refined},\mathbf{D}^s_{[1,t-1]}\right)$
\end{algorithmic}
\end{algorithm}






\section{Experiments}

\subsection{Datasets}
We evaluate on two long-term chat datasets, Multi-Session Chat(MSC)~\citep{xu2021beyond} and Conversation Chronicles (CC)~\citep{jang2023conversation}. While MSC is annotated by humans, CC is generated by ChatGPT. Both have 5 sessions of dialogue in each episode with various time intervals and persona. For testing, we use about 5 utterances per session across 50 episodes from each datasets, yielding approximately 250 sessions and 1250 utterances for each dataset.

\setlength{\extrarowheight}{3pt}
\begin{table}[hbt]
\small
\centering
\resizebox{0.8\textwidth}{!}{
\begin{tabular}{c|c|c|c|c}
\toprule

\textsc{} & $\text{SumMem}_\text{MSC}$ 
& $\textsc{ReBot}_\text{CC}$ & {$\text{GPT}_\text{CareCall}$} & \textsc{CREEM} \\ 
\hline
\textbf{What} & present& present & present &\makecell{present + past  } \\ 
\hline
\textbf{When} & per turn  & per session & per session & per turn \\ 
\hline
\textbf{Model} & \makecell{supervised \\encoder-decoder} & \makecell{distilled \\T5-base} &\makecell{ChatGPT}  & \makecell{ChatGPT} \\ 
\hline
\textbf{Fine-tuned}  & \makecell{$\checkmark$}& \makecell{$\checkmark$} & \makecell{$\times$} & \makecell{$\times$}\\ 

\bottomrule

\end{tabular}}
\caption{Comparison of memory constructing methods.}
\label{tab:model_compare_table}
\end{table}

\subsection{Models and Settings}
We compare our memory approach with the methods introduced with the datasets and another method reflecting the refining process of \citet{bae2022keep}.
\paragraph{$\text{SumMem}_\text{MSC}$} For the MSC dataset, they employ its own encoder-decoder abstractive summarizer, which we will call as $\text{SumMem}_\text{MSC}$, to extract persona for each speaker's utterance. 
\paragraph{$\textsc{ReBoT}_\text{CC}$}For the \textsc{CC} dataset, $\textsc{ReBoT}_\text{CC}$ summarizes each session's dialogue with a distilled T5-base model~\citep{raffel2020exploring}, using these summaries as long-term memory for both speakers. The T5-base model is trained with the summaries of 80K sessions generated by ChatGPT.

We set the two baseline methods to receive the entire memory as input for response generation, following the original $\textsc{ReBoT}_\text{CC}$ approach. \paragraph{$\text{GPT}_\text{CareCall}$}To evaluate our refining method, we experiment with the memory management mechanism proposed in \citet{bae2022keep}. Since there is no publicly available model, we implemented each operation using ChatGPT which we call $\text{GPT}_\text{CareCall}$. We implemented the methods using ChatGPT and Contriever to replicate the same model settings with our method for a fair comparison in the experiments. $\text{GPT}_\text{CareCall}$ performs pairwise processing of the existing memory and the summary generated at the end of the session, one by one; when memory already contains information of summary (PASS); when memory and summary are inconsistent, or when summary contains information of memory (REPLACE); when memory and summary are not related (APPEND); when memory is no longer true and summary is no longer needed to be remembered (DELETE).
\paragraph{\textsc{CREEM}} For \textsc{CREEM}, Contriever~\citep{izacard2021unsupervised} is used as the retriever to extract 5 information pieces each for question query and dialogue context query. Before each turn to respond, we instruct ChatGPT to generate 2 new insights to be added to the memory. We set the temperature parameter to 0.7 in order to create diverse insights. We provide detailed prompts in Appendix~\ref{app:CREEM_method} for reproducibility.

Table~\ref{tab:model_compare_table} provides a detailed comparison. For fair comparison regarding the memory systems, we use ChatGPT as the response generation model for all methods.

\subsection{Evaluation Metrics}
\subsubsection{Memory Evaluation}
Before assessing memory performance, it is essential to establish the criteria for what defines an effective memory. A good memory should consistently offer useful information aligned with past conversations. Similar to human memory, it should evolve, keeping track of changes and resolving conflicts between old and updated information. 
We establish three criteria for evaluating the qualities as a next-generation memory, under the \textit{evolutionary} aspects: 
\begin{itemize}
    \item \textbf{Integration}: accurate connection and reflection of changes across sessions in details
    \item \textbf{Consistency}: absence of contradictory or conflicting information
    \item \textbf{Sophistication}: advancement beyond one-dimensional, direct information
\end{itemize}

We set additional three criteria to assess the \textit{fundamental} aspects of a memory:
\begin{itemize}
    \item \textbf{Relevance}: closely related to the speakers
    \item \textbf{Concreteness}: includes specific and precise information
    \item \textbf{Diversity}: variety of topics, perspectives, or insights
\end{itemize}

We conduct G-Eval~\citep{liu2023gpteval} customized to evaluate the memory given each of the criteria, instruction steps, and the conversation of all sessions. 
\begin{table*}[t]
\small
\centering
 \resizebox{1.\textwidth}{!}{
 
\begin{tabular}{cccccccc}
\toprule 
\multirow{2}{*}{\textbf{Dataset}} & \multirow{2}{*}{\textbf{Model}} & 
  \multicolumn{3}{c}{\textbf{Evolutionary}} &
  \multicolumn{3}{c}{\textbf{Fundamental}} \\ 
  \cmidrule(lr){3-5} \cmidrule(l){6-8}
    
\multicolumn{2}{c}{} &Integration & Consistency & Sophistication  &  Relevance & Concreteness  &  Diversity\\
   \midrule

\multirow{4}{*}{MSC} &
 $\text{SumMem}_\text{MSC}$ & 3.44 & 3.95 & 3.02 & 4.09  &  3.82 & 3.72 \\  
& $\textsc{ReBot}_\text{CC}$ & 3.86 &  4.11 & 3.52  & 4.23  & 3.97 &  4.00 \\
& {$\text{GPT}_\text{CareCall}$} & 3.86 & 4.69 & 3.33  & 4.14 &  3.52 & 3.44 \\
& \textsc{CREEM} & \textbf{4.13} & \textbf{4.72} & \textbf{3.72} & \textbf{4.46}  &   \textbf{4.19} & \textbf{4.22}\\ 
\midrule 
 \multirow{4}{*}{CC} &
$\text{SumMem}_\text{MSC}$ &  3.16 & 4.11 & 2.59  &3.85  &  3.10 &  3.24 \\  
& $\textsc{ReBot}_\text{CC}$ & 4.01 & 4.50 & 3.57  & 4.51 &  4.01 & 4.00 \\
&{$\text{GPT}_\text{CareCall}$} & 3.88 & 4.87 & 3.67  & 4.30 &  3.59 & 3.21 \\
& \textsc{CREEM} & \textbf{4.17} & \textbf{4.94} & \textbf{3.83} & \textbf{4.61}  &  \textbf{4.19} & \textbf{4.03} \\ 

\bottomrule

\end{tabular}}
\caption{G-Eval of memory based on 6 criteria with CC and MSC dataset.}
\label{tab:mem_eval_results}
\end{table*}

\subsubsection{Contradiction Level Evaluation}
To conduct a more comprehensive evaluation focusing on the contradictory aspect, we assess the contradiction level by computing the average probability of contradiction for all pairs within the memory. This evaluation is performed using RoBERTa~\citep{liu2019roberta}, which is fine-tuned on the MultiNLI(MNLI) dataset~\citep{williams2017broad}.

\subsubsection{Question-answering Evaluation}
To conduct a more robust evaluation of the refining process, we performed a question-and-answering(QA) based evaluation for \textsc{CREEM} and $\text{GPT}_\text{CareCall}$. 
\paragraph{Question Generation}There have been studies of evaluating abstractive summarization through QA-based evaluation~\citep{scialom2019answers, wang2020asking}. 
We create questions specifically designed to assess the memory system's ability to remember altered information throughout the session.
Utilizing the Chain-of-Thoughts~\citep{wei2022chain} and form-filling technique, we instruct ChatGPT to generated multiple-choice questions.
Given the entire session dialogue, ChatGPT identifies the changed information denoted as [Before] and [After]. Then, it creats questions based on this captured information. This process ensures that shifts in the speaker's status to be accurately reflected within the problem. 
Additionally, a "No Answer" option is included for cases where the question cannot be answered based on its memory. Please refer to the Appendix~\ref{app:Question_generating} for detailed prompts. 
\paragraph{Answering}
The quality of memory is evaluated based on how well an answering model utilized the given memory to answer questions, measured by the accuracy rate of its responses. We utilize ChatGPT as the answering model, providing each memory as the sole external information. 

\subsubsection{Response Generation Evaluation}
It is necessary to verify whether the quality of memory aligns with the quality of responses. To evaluate response quality, we conduct reference evaluations using BLEU, ROUGE-L, and $\textsc{BERTScore}$~\citep{zhang2019bertscore,
papineni2002bleu, lin2004rouge}. However, a high score in these metrics does not necessarily imply that the model generated a superior response. Given the potential diversity in responses even within the same context, deviation from the reference response should not be interpreted as an incorrect answer. 

Hence, alongside the automated evaluations, we perform a reference-free pair-wise evaluation using ChatGPT. 
Since we are focusing on the memory's impact rather than the response generation performance itself, we established 2 criteria that can most effectively show the role of the memory: 
\begin{itemize}
    \item \textbf{Consistency}: maintains consistency without contradictions from past sessions
    \item \textbf{Memorability}: effectively recalls past events correctly by retaining information from previous sessions
\end{itemize}
\begin{figure}[t]
\begin{minipage}{\textwidth}
\begin{minipage}[b]{0.51\textwidth}
    \centering
    \resizebox{1.1\textwidth}{!}{
    \begin{tabular}{cccccc}
\toprule 
\multirow{2}{*}{\textbf{Dataset}} & \multirow{2}{*}{\textbf{Model}} & \multicolumn{4}{c}{Session}\\
\cmidrule(l){3-6}
\multicolumn{2}{c}{} & 2 & 3 & 4 & 5 \\  
\midrule
\multirow{4}{*}{MSC} & $\text{SumMem}_\text{MSC}$ & 0.3475 & 0.3514 & 0.363 & 0.3658 \\  
& $\textsc{ReBot}_\text{CC}$ & 0.198 &  0.2065 & 0.2157 & 0.2228 \\
& {$\text{GPT}_\text{CareCall}$} & 0.3054 & 0.2653 & 0.2916 & 0.2678  \\
    & \textsc{CREEM} & \textbf{0.1493} & \textbf{0.1476} & \textbf{0.1539} & \textbf{0.1613} \\ 
 \midrule 
\multirow{4}{*}{CC} & $\text{SumMem}_\text{MSC}$ &  0.1476 & 0.1656 & 0.1743 & 0.1775  \\  
& $\textsc{ReBot}_\text{CC}$ &  0.1717 & 0.177 & 0.2054 & 0.2006  \\  
&  {$\text{GPT}_\text{CareCall}$} & 0.1176 & \textbf{0.1036} & \textbf{0.1583} & \textbf{0.1241}\\
& \textsc{CREEM} & \textbf{0.11} & 0.1454 & 0.1872 & 0.1915  \\ 
\bottomrule
    \end{tabular}}
\captionof{table}{Degree of Contradiction.}\label{tab:contradictory_NLI}
\end{minipage}
  \hfill
    \begin{minipage}[b]{0.47\textwidth}
    \centering
    \begin{center}
        \includegraphics[width=0.8\linewidth]{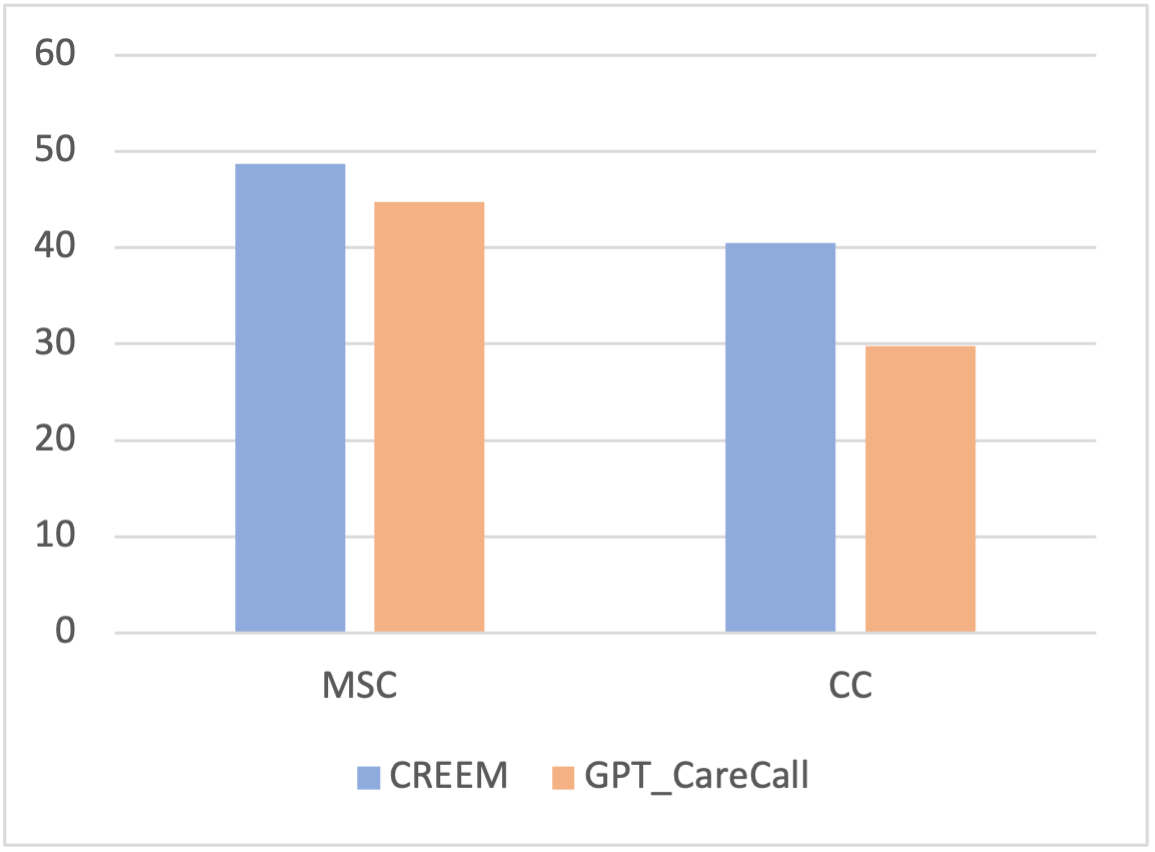}
    \end{center}
    \captionof{figure}{QA Accuracy Rate.}\label{QA_correction_chart}
    \end{minipage}
  \end{minipage}
\end{figure}

 




\section{Results}
\subsection{Main Results}
\paragraph{\textsc{CREEM} makes high quality memory.}

Table~\ref{tab:mem_eval_results} displays scores of G-eval over memory quality. \textsc{CREEM} outperforms in all three evolutionary aspects, indicating its superior ability to integrate, maintain consistency, achieve sophistication. The \textbf{Blend} process appears to play a crucial role in integrating information across sessions, creating sophisticated memories by expanding the memory. Additionally, the \textbf{Refine} process contributes to maintaining a consistent and well-organized memory by continually removing redundant or outdated information.
Even in fundamental aspects, \textsc{CREEM} excelled in all criteria.

\paragraph{\textsc{CREEM}'s refining operations are effective in maintaining memory consistency.}
Table~\ref{tab:contradictory_NLI} presents the degree of contradiction within the memory predicted by the NLI model. 
Overall, \textsc{CREEM} exhibited a low level of contradiction. However, concerning the CC dataset, {$\text{GPT}_\text{CareCall}$} showed even lower levels of contradiction. While \textsc{CREEM} merge contradictory information into one, {$\text{GPT}_\text{CareCall}$} remove both or one of the contradictory information through its operations, resulting in a higher removal of contradictory data. This approach may lead to excessive information loss, as evidenced by QA evaluation results in Figure~\ref{QA_correction_chart}. \textsc{CREEM} shows higher accuracy rates in both datasets. The lower accuracy rates for {$\text{GPT}_\text{CareCall}$} suggest that it may have missed significant information during the refinement process. These results show that \textsc{CREEM}'s \textbf{Blend} and \textbf{Refine} process effectively incorporates both past and future information, maintaining better memory retention while reducing contradiction.
\begin{table*}[t]
\small
\centering
 \renewcommand{\arraystretch}{1.1}
\begin{tabular}{ccccccc}
\toprule 
\multirow{2}{*}{Model} & \multicolumn{3}{c}{MSC} & \multicolumn{3}{c}{CC}\\ 
\cmidrule(lr){2-4} \cmidrule(l){5-7}

& BLEU & ROUGE-L & $\textsc{BERTScore}$ & BLEU & ROUGE-L & $\textsc{BERTScore}$\\
\midrule
$\text{SumMem}_\text{MSC}$ &  1.392 & 0.143 & \textbf{0.866} & 4.053 & 0.175 & 0.876  \\  
$\textsc{ReBot}_\text{CC}$ &  1.416 & 0.148 & 0.863 & 3.861 & 0.195 & 0.877  \\  
{$\text{GPT}_\text{CareCall}$} & 1.485 & 0.139 & 0.862 & 3.987 & 0.171 & 0.873 \\

\textsc{CREEM} & \textbf{1.610} & \textbf{0.150} & 0.865 &\textbf{ 4.667} & \textbf{0.198} & \textbf{0.880}  \\ 

\bottomrule
\end{tabular}
\caption{Performance in response generation by BLEU, ROUGE-L, and \textsc{BERTScore}}
\label{tab:reference_scoring}
\end{table*}
\begin{table*}[t]
 \resizebox{\columnwidth}{!}{
\renewcommand{\arraystretch}{1.3}
\begin{tabular}{cccccccc}
\toprule

\multirow{2}{*}{\textbf{Dataset}} & \multirow{2}{*}{\textbf{Criteria}} & \multicolumn{6}{c}{\textbf{ \textsc{CREEM}} vs.} \\
\cmidrule(lr){3-8}

& & \textsc{ $\textsc{ReBoT}_\text{CC}$} &
{ $\text{SumMem}_\text{MSC}$} & {$\text{GPT}_\text{CareCall}$} & $-$ Question & $-$ Refine &
$-$ Insight Infuse \\  
\hline
\multirow{2}{*}{MSC} & Consistency & \textbf{ 81.52} \% & \textbf{ 60.09} \%  & \textbf{76.45} \%  & \textbf{62.67} \% & \textbf{72.88 \%} & \textbf{ 72.37 \%} \\
& Memorability & \textbf{ 71.01} \% & \textbf{ 50.00} \%  & \textbf{68.12} \%  & \textbf{73.63} \% & \textbf{ 70.34 \%} & \textbf{ 69.30 \%}\\
\hline
\multirow{2}{*}{CC} & Consistency & \textbf{ 76.45} \% & \textbf{ 69.12} \% & \textbf{81.29} \% & \textbf{ 66.21} \% & \textbf{ 59.26} \% & \textbf{ 67.53} \%   \\
& Memorability & \textbf{ 54.52} \% & \textbf{ 56.87} \% & \textbf{72.58} \% & 49.68 \% & \textbf{ 58.02} \% & \textbf{ 54.32 \%}\\

\bottomrule

\end{tabular}}
\caption{Comparison of generated response quality. We report \textsc{CREEM}’s winning rate.}
\label{tab:rg_winrate_table}
\end{table*}

\paragraph{\textsc{CREEM}'s memory improves response generation.}

Table~\ref{tab:reference_scoring} presents the reference-based evaluation of the quality of responses generated by each model across different datasets. Under the \textsc{BERTScore} metric, the MSC data shows slightly better results with its own fine-tuned model, $\text{SumMem}_\text{MSC}$. Aside from that, it shows that \textsc{CREEM} generates better responses even without fine-tuning. 
Table~\ref{tab:rg_winrate_table} shows the winning rate of \textsc{CREEM}. Overall, ChatGPT generates better answers when utilizing \textsc{CREEM}'s memory, compared to three other memories. Particularly in terms of consistency, ChatGPT receives better evaluations, indicating that the \textbf{Refine} process effectively generates consistent responses aligned with past conversations. We can conclude that the performance of the response generation is positively influenced by the quality of the memory.

\subsection{Ablation Study}
\paragraph{Contextual searching using generated questions is effective for response quality.}
Table ~\ref{tab:rg_winrate_table} shows the performance difference between \textsc{CREEM} and its variations, excluding question generation and only using the current dialogue as the query in the \textbf{Contextual Search} process. 
Except for the memorablity in CC data, the original version of including both question and dialogue as query tends to make better responses. This show that searching with the generated question retrieves a wider range of information, which can contribute to creating a better memory.

\paragraph{Refinement has a crucial effect both in memory and response quality.}
Table ~\ref{tab:rg_winrate_table} shows excluding the \textbf{Refine} process degrades the response quality. 
It can be observed that refining leads to better memory retention and consistent responses compared to simply accumulating all information without refinement.

\paragraph{Pre-response blending helps generate better response.}
Table~\ref{tab:rg_winrate_table} shows the performance difference between \textsc{CREEM} and its variants excluding the \textbf{Insight Infusing} process right before generating a response. This setting only gives memories accumulated in the past session, while \textsc{CREEM}'s original method also provides new memories created in the ongoing session until its responding turn. 
Overall, \textsc{CREEM} consistently produces more consistent and memory-enhanced responses. 

\begin{figure*}[t] 
    \begin{center}
    \includegraphics[width=1.0\linewidth]{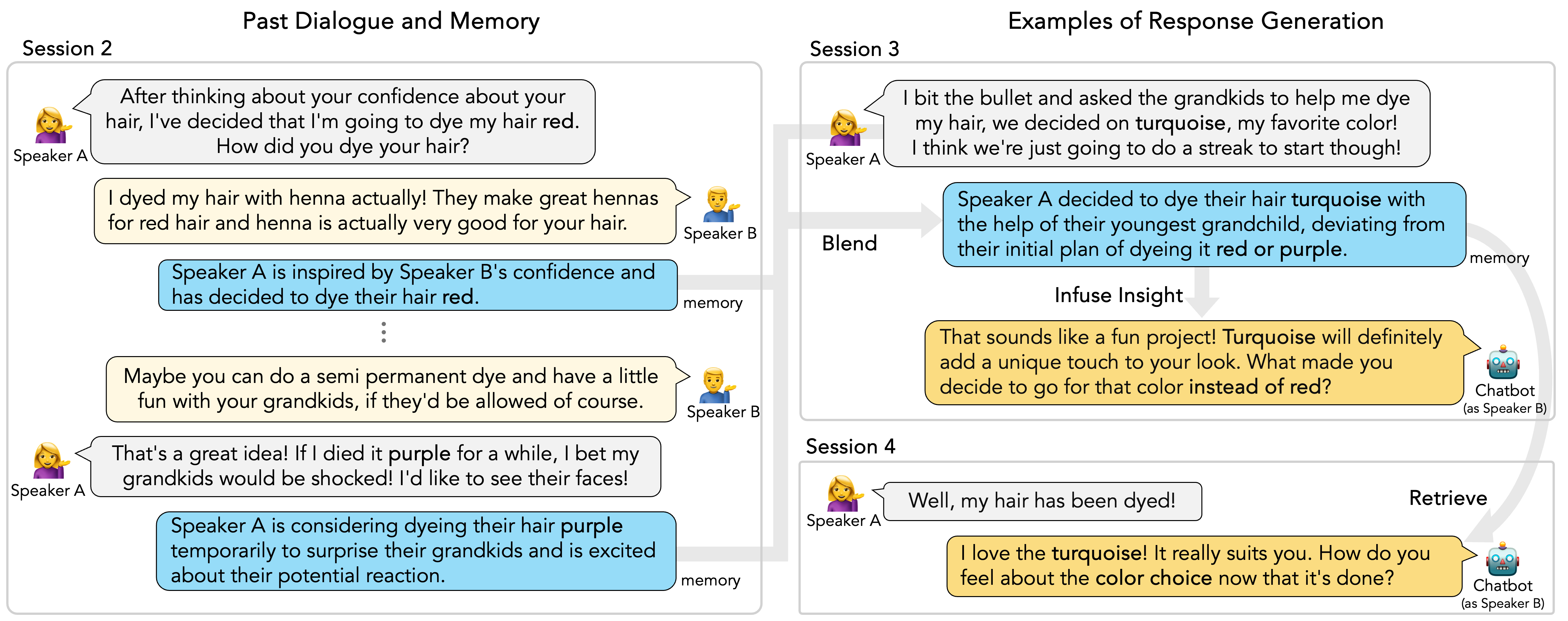}
    \end{center}
    \caption{Example episode of how past memory is blended and utilized. Change in Speaker A’s hair color choice is reflected in the later sessions.
}
    \label{Overall Method Example}
\end{figure*} 
\section{Conclusion}


We create an ever-evolving memory scheme to properly integrate both present and past memories, actively refining outdated or contradictory information. Despite using a maximum 2-shot LLM prompting approach, our method, \textsc{CREEM}, outperforms other fine-tuned methods, $\text{SumMem}_\text{MSC}$ and $\textsc{ReBot}_\text{CC}$, in terms of memory alignment and response generation. 
Even when compared with with the $\text{GPT}_\text{CareCall}$ which applies different refining methods within the same settings with \textsc{CREEM} using ChatGPT, we have demonstrated the effectiveness of our refining method in improving memory retention and consequently enhancing the quality of responses. 
Through \textsc{CREEM}, we aim to create a more human-like memory system, fostering further evolution of chat-bot memory systems.

\section{Limitation}

Our study has the following limitations: (1) Apart from the scheme of \textsc{CREEM}, the results can be influenced by the performance of the retriever. As future work, we plan to develop a retriever model specialized in memory and conversation data;
(2) During the \textbf{Blend} \& \textbf{Refine} process, CREEM operates within the top-k retrieved memories. However, it may overlook removing insights or updated relationships when the disappearing past memories are less than or equal to k. Currently, our method doesn't handle unretrieved memories, but incorporating graph relationships among memories in the future could enhance memory retrieval and management;
(3) There is not enough publicly open high-quality long-term datasets. We expect \textsc{CREEM} to have an advantage over other models as the conversations become very long through the blend and refine process. We plan to experiment by developing a longer dataset for better memory evaluation.

\bibliography{reference,custom}
\bibliographystyle{colm2024_conference}
\newpage

\appendix
\section{Method Prompts}
\label{app:CREEM_method}
\subsection{Query Prompt}
We use ChatGPT to create an appropriate query that could be generated for all conversations from the current session. We use a two-shot prompt for this task, but we don't include all examples due to the length. The prompt for this task is as follows: \\
{\fontfamily{qcr}\selectfont
You will act as a model that generates a query relevant to the given dialogue. \\
Given the dialogue, what is the most salient high-level question we can answer about {speaker}, focusing mainly on the last utterance?\\
... \\
Dialogue:\{text\} \\
Question about \{speaker\}:
}

\subsection{Blend for first session Prompt}
We use ChatGPT to generate two insights for the first session. For this task, a one-shot prompt is used, and examples are omitted for length. The prompt is as follows:\\
{\fontfamily{qcr}\selectfont 
Create insights by extracting detailed information from the <Current Dialogue>.\\
Here is an example that can be helpful.\\
... \\
\text{[INPUT]}\\
<Current Dialogue>\\
\{current\_dialogue\} \\
\\
\text{[OUTPUT]}
}

\subsection{Blend and Refine Prompt}
We use ChatGPT to generate two insights related to insights and utterances that have been retrieved for queries in a second or later session. For this task, a two-shot prompt is used, and examples are omitted for length. The prompt is as follows:\\
{\fontfamily{qcr}\selectfont 
Create insights by extracting detailed information from the LAST TWO utterences of the <Current Dialogue> and reflecting the <Past Memories>.\\
You can make three types of insights. In one insight, include information about only one speaker.\\
1. \text{[Updated]}\\
If any information from the <Past Memories> is changed in the <Current Dialogue>, create an Insight reflecting the changes.\\
Only update the memory with the corresponding speaker's new information.\\
Then, indicate which past memory has been changed to remove.\\
2. \text{[Redundant]}\\
If <Current Dialogue> contains redundant information of the <Past Memories>, merge them including the details.\\
Then, indicate which past memory is redundant to remove.\\
3. \text{[New]}\\
If there is new information that is not related to the <Past Memories>, create a new Insight.\\
Then, write "None" and do not remove anything.\\
\\
Here are some an examples.\\
... \\
\text{[INPUT]}\\
<Past Memories>\\
\{retrieved\_memories\}\\
\\
<Current Dialogue>:\\
\{current\_dialogue\} \\
\\
\text{[OUTPUT]}
}

\subsection{Respond Prompt}
After the Refine stage, ChatGPT  generate an appropriate response to the previous utterance based on the newly established memory. Three guidelines are given to create an appropriate response. The prompt is as follows:\\
{\fontfamily{qcr}\selectfont
You will act as a response generator. Considering the given [Memory], continue the given conversation and create an appropriate response for the last utterance.\\
Consider the following factors when generating:\\
  - Do not make the response too long.\\
  - Do not contain any of the speaker's name explicitly.\\
  - Reflect the [Memory] in the generated response.\\
...\\
\text{[Memory of \{speaker\}]}:\\
\{memory\_stream\}:\\
\text{[Conversation]}:\\
\{current\_dialogue\}:
\text{[Response] \{speaker\}}: 
}
\section{Evaluation Prompts}
\subsection{Memory Evaluation Prompt}
\label{app:memory_evaluation}
We compare which of the memories created by $\text{SumMem}_\text{MSC}$, $\textsc{ReBoT}_\text{CC}$, {$\text{GPT}_\text{CareCall}$}, and \textsc{CREEM} had better performance through ChatGPT using the CC dataset and the MSC dataset. There are a total of 6 criteria used in this evaluation, and each prompt contains different evaluation steps for each criteria so that we can evaluate which memory better reflects the conversation by looking at the total dialogue and the memory created for the dialogue. Considering that the results are slightly different each time ChatGPT generates, a total of 20 results are generated at a time and the average is taken as the result. The evaluation step is omitted due to length. The seven criteria are as follows:\\
\begin{itemize}
    \item Integration (1-5): This criterion evaluates whether the memory effectively connects information across sessions and accurately reflects changes in details about \{speaker\} over time. The goal is to prioritize memories that present a cohesive organization of content across sessions.
    \item Consistency (1-5): This criterion evaluates whether there are any contradictory or conflicting pieces of information about \{speaker\} within the memory. The goal is to ensure that the information presented about \{speaker\} remains coherent and free from contradictions throughout the conversation.
    \item Sophistication (1-5): This criterion assesses whether the content of the memory is advanced and goes beyond one-dimensional, direct information about \{speaker\} from the conversation. 
    \item Relevance (1-5): This criterion evaluates whether the memory is closely related to the \{speaker\} of the conversation. It ensures that the content contributes meaningfully to the ongoing conversation.
    \item Concreteness (1-5): This criterion evaluates whether the memory contains substantial and specific information about \{speaker\}. It assesses the level of detail and precision in the representation of \{speaker\}.
    \item Diversity (1-5): This criterion assesses whether the content of the memory has diverse information about \{speaker\}.
\end{itemize}
\subsection{Question Generating Prompt}
\label{app:Question_generating}
We use ChatGPT to view the entire conversation session and create multiple-choice questions that captured the changes. The prompts used to create the problem are as follows, and examples of prompts are omitted for length:
{\fontfamily{qcr}\selectfont
Your task is to create three questions multiple choices that can be answered from the given dialogue. Actively reference the contents of the dialogue in the options, but try to use different expressions. \\
You will make several questions asking only about the target speaker. The subject of the question should be the target speaker.\\
Focus on the changed status of the target speaker and use that as a source for making questions.\\
To make the question difficult, you should follow these steps:\\
1.  Find any information that has changed about the target speaker during the conversation.\\
The format should be: "[Before] past information [After] newly updated information"\\
2. Generate questions that asks about the changed information.\\
3. Next step is to make multiple answer options that is confusing. Make a correct choice from the newly updated information mentioned after the [After] token, and wrong choices from past outdated information mentioned after the [Before] token.\\
4. Write the correct answer with a brief explanation.\\
\\
Here is an example question.\\
...\\
Now generate 3 questions about the target speaker, \{target\_speaker\}.\\
Target Speaker: \{target\_speaker\} \\
\text{[Dialogue]}\\
\{dialogue\} \\
\text{[Information about \{target\_speaker\}]}
}

\subsection{Question Answering Prompt}
\label{app:Question_Answering}
We use ChatGPT to solve the problems created through the above prompt through the memories of each model. The prompt to solve the problem using memories is as follows:
{\fontfamily{qcr}\selectfont
Based on the given memory, solve the three questions by choosing one of the options to answer the following question. Also choose the sentences that support your answer.\\
Make sure to answer based on ONLY the given memory, without prior knowledge.\\
The answer might not be directly addressed in the memory so try hard to infer the most reasonable answer among choices A, B, C. But if you still cannot find any answers using the memory, choose "D) No answer".\\
\\
The answer format should be:\\
\\
\text{[Answers]}:\\
Answer 1: \\
A) She is a very active and lively.\\
\\
Supporting sentences: \\
Sentence 2) She has three kids and her mom is pediatrician. \\
Sentence 5) She commutes by boat every day to work and lives in amsterdam with canals.\\
\\
Now solve the questions.\\
\\
\text{[Memory]}:\\
\\
\{memory\}\\
\\
\text{[Questions about Co-worker B]} \\
\\
\{questions\} \\
\\
\text{[Answers]}:\\
}

\subsection{Response Evaluation Prompt}
\label{app:response_evaluation}
We use ChatGPT to pairwise compare the responses generated by \textsc{CREEM} with the responses generated by $\text{SumMem}_\text{MSC}$, $\textsc{ReBoT}_\text{CC}$ and {$\text{GPT}_\text{CareCall}$}. In conducting this work, two criteria are used:
consistency and memorability. An example of the format is given and the criteria were specified. The criteria according to the evaluation criteria are as follows. 
\begin{itemize}
    \item Consistency: The response should maintain consistency without contradictions from past sessions. The response aligns seamlessly with what has been discussed in previous sessions, following the individual's characteristic and current status.
    \item Memorability: The response should effectively recalls past events correctly by retaining information from previous sessions. The response actively uses memory from the past dialogue to continue the conversation.
\end{itemize}
Considering that the results are slightly different each time ChatGPT generates, a total of 19 results were generated at once, and the one selected with more was taken as the result. Among the two prompts, the following is one that evaluates consistency. Format examples have been omitted due to length:\\
{\fontfamily{qcr}\selectfont
You will be given a conversation between two individuals. You will then be given two response options for the next turn in the conversation.\\
Your task is to choose the better response based on one metric with a brief explanation.\\
Please make sure you read and understand these instructions carefully. Please keep this document open while reviewing, and refer to it as needed.\\
\\
Evaluation Criteria:\\
\\
Consistency - The response should maintain consistency without contradictions from past sessions. The response aligns seamlessly with what has been discussed in previous sessions, following the individual's characteristic and current status.\\
...\\
Now choose the response that has better consistency given the past and current session dialogue.\\ 
\\
Past Sessions:\\
\{past\_dialogue\}\\
\\
Current Session:\\
\{current\_dialogue\}\\
\\
Response Options:\\
\{response\}\\
\\
Better Reponse:\\
}

\section{Examples of Memory construction and Response generation}
Table~\ref{tab:comparison of memory and response generation} shows an example of building a memory and generating an answer using the methods of $\text{SumMem}_\text{MSC}$, $\textsc{ReBoT}_\text{CC}$, and \textsc{CREEM} for one episode from a random MSC dataset.
\begin{table}[t]
\centering
\resizebox{0.7\textwidth}{!}{
\begin{tabular}{c|ccc}
\toprule
& \textbf{\textit{Redundant}} & \textbf{\textit{New}} & \textbf{\textit{Updated}}\\
\midrule
\textbf{MSC dataset} & 24.61 \% & 55.81 \% & 19.59 \%\\
\textbf{CC dataset} & 30.35 \% & 53.06 \% & 16.59 \%\\

\bottomrule
\end{tabular}}
\caption{Ratio of \textit{Redundant}, \textit{New}, \textit{Updated} during constructing memory by \textsc{CREEM}} 
\label{tab:ratio of redundant, new, updated}
\end{table}

\begin{table*}[t]
\small
\centering
 \resizebox{1.\textwidth}{!}{
 
\begin{tabular}{lll}
\toprule 
  \multicolumn{3}{l}{Dialogue Context (\textit{msc-episode-40-s4-t5})} \\
  \midrule
  \multicolumn{3}{l}{Speaker A: Well, my hair has been dyed!} \\
  \multicolumn{3}{l}{Speaker B: How did it come out?  I bet your grandkids were thrilled.lol}\\
  \multicolumn{3}{l}{Speaker A: Oh my youngest did so good! You can tell that being a}\\
  \multicolumn{3}{l}{cosmetologist is her passion, she was an absolute natural with keeping my nerves at bay. }\\ 
  \multicolumn{3}{l}{Speaker B: At least it's less permanent than a drastic hair cut would be.}\\
  \multicolumn{3}{l}{You've got extra flavor now. Did you stick with just a streak or did you go bigger?}\\
  \multicolumn{3}{l}{Speaker A: We decided to do a streak on each side, for balance. And they're }\\
  \multicolumn{3}{l}{not just out in the open but under a layer of hair, so it pops out when I wear }\\
  \multicolumn{3}{l}{a ponytail. My granddaughter suggested that, how creative!}\\
   \midrule

$\text{SumMem}_\text{MSC}$ \\
\hspace{0.3cm} Memory & \multicolumn{2}{l}{I am excited for Easter Weekend. I have blue hair. I have a new Easter dress.}\\
& \multicolumn{2}{l}{I am 29 years old. I am going on spring break. My mom just gave me a new car. I drive.}\\
& \multicolumn{2}{l}{I dyed my hair with henna. I love grey hair. I want to dye my hair grey. My dress is turquoise.}\\
& \multicolumn{2}{l}{My hair is blue. I just got my first new car. I've been driving pre owned cars before.}\\
& \multicolumn{2}{l}{I have red hair. I have a friend. I have a VW Tiguan. I fill up my car once a month.}\\
 & \multicolumn{2}{l}{I don't drive a lot. I work as a librarian at the beach.}\\
\cmidrule(lr){2-3}
\hspace{0.3cm} Response & \multicolumn{2}{l}{That sounds like a fun and unique way to add some color to your hair!}\\
&\multicolumn{2}{l}{It's great that your granddaughter was able to offer a creative suggestion. I'm sure it looks amazing}\\
&\multicolumn{2}{l}{when you wear a ponytail.}\\
 \midrule
\textsc{$\textsc{ReBoT}_\text{CC}$} \\
\hspace{0.3cm} Memory & \multicolumn{2}{l}{A retired librarian plans to leave for spring break due to her mother's new car }\\
& \multicolumn{2}{l}{and quiet nature. She enjoys nature, but is retired due to retiring.}\\
& \multicolumn{2}{l}{A woman dyes her hair red with henna, which is good for red hair. She also suggests }\\
& \multicolumn{2}{l}{doing a semi-permanent dye and having fun with her grandkids if allowed . }\\
& \multicolumn{2}{l}{Turquoise is her favorite color, but her new dress is turquoise.}\\
& \multicolumn{2}{l}{A grandkid dyed her hair turquoise, which she thinks will look great on her youngest }\\
& \multicolumn{2}{l}{grandson's favorite color. They plan to touch up their hair for the summer months due to }\\
& \multicolumn{2}{l}{the sun and pool. They also have recommendations for a day visit at the beach."}\\
\cmidrule(lr){2-3} 
\hspace{0.3cm} Response & \multicolumn{2}{l}{That sounds like a fun and unique way to add some color to your hair! I'm glad your}\\
& \multicolumn{2}{l}{granddaughter had such a creative idea. It must be exciting to have a little pop of}\\
& \multicolumn{2}{l}{ color that shows when you wear a ponytail.}\\
\midrule
{$\text{GPT}_\text{CareCall}$}\\
\hspace{0.3cm} Memory & \multicolumn{2}{l}{Speaker A has asked their youngest grandkid to help them dye their hair turquoise,}\\
& \multicolumn{2}{l}{as they want to get into hairdressing as a profession.}\\
& \multicolumn{2}{l}{Speaker A jokes that nature will eventually give Speaker B grey hair.}\\
& \multicolumn{2}{l}{Speaker A recommended the book \"Beautiful Ruins\" by Jess Walter to Speaker B}\\
& \multicolumn{2}{l}{which is a fiction book about Hollywood and actors/actresses."}\\
& \multicolumn{2}{l}{Speaker B recently bought a VW Tiguan in a red color.}\\
\cmidrule(lr){2-3}
 \hspace{0.3cm} Response 
 & \multicolumn{2}{l}{And speaking of hair, I remembered you just got that red VW Tiguan.}\\
 & \multicolumn{2}{l}{So now we both have a pop of color in our lives!}\\
\midrule
 \textsc{CREEM}\\
 \hspace{0.3cm} Memory & \multicolumn{2}{l}{Speaker B is open to experimenting with different hair colors and is specifically interested}\\
 & \multicolumn{2}{l}{in dyeing their hair grey in the future. Speaker B highly values the opinions and}\\
 & \multicolumn{2}{l}{expertise of their grandkids when it comes to hair dyeing decisions.}\\
 \cmidrule(lr){2-3}
 \hspace{0.3cm} Response & \multicolumn{2}{l}{Your grandkids really have a good eye for hair styling! I remember you mentioning that}\\
 & \multicolumn{2}{l}{you were interested in dyeing your hair grey in the future. Maybe you can consult}\\
 & \multicolumn{2}{l}{ them for some ideas when the time comes.}\\

\bottomrule

\end{tabular}}
\caption{Comparison of memory and response generation between \textsc{CREEM}, $\text{SumMem}_\text{MSC}$, \textsc{$\textsc{ReBoT}_\text{CC}$}  for MSC dataset}
\label{tab:comparison of memory and response generation}
\end{table*}


\section{Analysis of \textbf{Refine}}
Table~\ref{tab:ratio of redundant, new, updated} shows the extent to which \textit{Redundant}, \textit{New}, and \textit{Updated} processes occur when creating memory with \text{CREEM} through 50 MSC datasets and 50 CC datasets. When using the MSC dataset, the rate of \textit{Redundant} and \textit{Updated} occurrences is about 44.2 \%, and when using the CC dataset, it is about 46.94 \%. Thus, we confirm that the process of checking and refining whether there are redundant or updated memories among past memories was performed properly.

\section{Implementation details}
\subsection{Contriever}
In our implementation into \textsc{CREEM}, We use Contriever ~\cite{izacard2021unsupervised} to retrieve relevant past insights or utterances for queries created after the previous utterance. Contriever, trained through unsupervised contrastive learning, serves as a dense information retriever. Remarkably, even in the absence of explicit supervision, it exhibits outstanding proficiency in information retrieval tasks.
\subsection{Large Language Model}
In this study, we use ChatGPT for implementing Query Generation, Blend and Refine and Response Generation. ChatGPT, a Large Language Model (LLM) with 175 billion parameters, is built upon Instruct-GPT~\cite{ouyang2022training}. It is specifically trained to adhere to user instructions and provide requested information in a conversational style. To interact with the OpenAI API, we utilize LangChain.

\section{License and Terms}
For our implementation and evaluation, we use Huggingface library\footnote{\url{https://huggingface.co/}}.
This library licensed under Apache License, Version 2.0.
We have confirmed that all of the artifacts used in this paper are available for non-commercial scientific use.

\end{document}